\def\year{2022}\relax
\documentclass[letterpaper]{article} 
\usepackage{aaai22}  
\usepackage{times}  
\usepackage{helvet}  
\usepackage{courier}  
\usepackage[hyphens]{url}  
\usepackage{graphicx} 
\usepackage{natbib}  
\usepackage{caption} 
\DeclareCaptionStyle{ruled}{labelfont=normalfont,labelsep=colon,strut=off} 
\usepackage{algorithm}
\usepackage{algorithmic}

\usepackage{newfloat}
\usepackage{listings}
\floatstyle{ruled}
\newfloat{listing}{tb}{lst}{}
\floatname{listing}{Listing}
\usepackage{amsmath, amssymb}
\usepackage{enumitem}
\usepackage{color}
\newcommand{\magenta}[1]{\textcolor{magenta}{#1}}
\newcommand{\blue}[1]{\textcolor{blue}{#1}}

\usepackage{xspace}
\usepackage{multirow}
\usepackage{hhline}
\usepackage{colortbl}
\usepackage{soul}
\usepackage{bm}
\usepackage{comment}
\usepackage{balance}

\newenvironment{myprotocol}[1][!t]{%
    \floatname{algorithm}{Protocol}
   \begin{algorithm}[#1]%
   
  }{\end{algorithm}}
  
  \newcommand{\vecx}{\mathbf{x}}
\newcommand{\peqopp}{\ensuremath{\pi_{\mathsf{EOP}}}\xspace}
\newcommand{\boldpeqopp}{\ensuremath{\bm{\pi_{\mathsf{EOP}}}}\xspace}
\newcommand{\peqodd}{\ensuremath{\pi_{\mathsf{EOD}}}\xspace}
\newcommand{\boldpeqodd}{\ensuremath{\bm{\pi_{\mathsf{EOD}}}}\xspace}
\newcommand{\pdp}{\ensuremath{\pi_{\mathsf{DP}}}\xspace}

\newcommand{\pacc}{\ensuremath{\pi_{\mathsf{GACC}}}\xspace}
\newcommand{\boldpdp}{\ensuremath{\bm{\pi_{\mathsf{DP}}}}\xspace}
\newcommand{\boldpacc}{\ensuremath{\bm{\pi_{\mathsf{GACC}}}}\xspace}

\newcommand{\pinfer}{\ensuremath{\pi_{\mathsf{INFER}}}\xspace}
\newcommand{\peq}{\ensuremath{\pi_{\mathsf{EQ}}}\xspace}

\newcommand{\pmul}{\ensuremath{\pi_{\mathsf{MUL}}}\xspace}
\newcommand{\pdiv}{\ensuremath{\pi_{\mathsf{DIV}}}\xspace}

\newcommand{\indep}{\perp \!\!\! \perp}

\title{PrivFair: a Library for Privacy-Preserving Fairness Auditing}
\author{
    Sikha Pentyala,\textsuperscript{\rm 1} 
    David Melanson,\textsuperscript{\rm 2}
    Martine De Cock,\textsuperscript{\rm 2,3}
    Golnoosh Farnadi\textsuperscript{\rm 1,4}
}
\affiliations{
    \textsuperscript{\rm 1} Mila -- Quebec AI Institute\\
    \textsuperscript{\rm 2} School of Engineering and Technology, University of Washington Tacoma\\
    \textsuperscript{\rm 3} Department of Applied Mathematics, Computer Science and Statistics, Ghent University\\
    \textsuperscript{\rm 4} Department of Decision Sciences, HEC Montr\'eal\\
}

\newcommand{\golnoosh}[1]{\magenta{\textsc{Golnoosh:} #1}}

\newcommand{\sikha}[1]{\blue{\textsc{Sikha:} #1}}

\usepackage{bibentry}

\begin{document}

\maketitle

\begin{abstract}
Machine learning (ML) has become prominent in applications that directly affect people's quality of life, including in healthcare, justice, and finance. 
ML models have been found to exhibit discrimination based on sensitive attributes such as gender, race, or disability. Assessing if an ML model is free of bias remains challenging to date, and by definition has to be done with sensitive user characteristics that are subject of anti-discrimination and data protection law. Existing libraries for fairness auditing of ML models offer no mechanism to protect the privacy of the audit data.
We present \textsc{PrivFair}, a library for privacy-preserving fairness audits of ML models. Through the use of Secure Multiparty Computation (MPC), \textsc{PrivFair} protects the confidentiality of the model under audit and the sensitive data used for the audit, hence it supports scenarios in which a proprietary classifier owned by a company is audited using sensitive audit data from an external investigator. We demonstrate the use of \textsc{PrivFair} for group fairness auditing with tabular data or image data,
without requiring the investigator to disclose their data to anyone in an unencrypted manner, or the model owner to reveal their model parameters to anyone in plaintext.

\end{abstract}

\section{Introduction}
Algorithmic decision making, driven by machine learning (ML), has become very prominent in applications that directly affect people's quality of life, including in healthcare, justice, and finance.
ML models have made discriminatory inferences in recidivism prediction \cite{angwin2016machine}, credit card approval \cite{vigdor2019apple}, advertising \cite{ali2019discrimination} and job matching \cite{linkedin:2021}, among others. Concerns over fairness in ML have prompted research into the establishment of fairness metrics and techniques to mitigate bias; see e.g.~\cite{calders2010three,galhotra2017fairness,kearns2018preventing,pleiss2017fairness,verma2018fairness} and references therein. 
In state-of-the-art approaches based on \textit{group fairness measures}, 
audits are done by comparing the ML model predictions for different demographic subgroups in the audit data set. For instance,  a classifier satisfies the definition of \textit{demographic parity} \cite{dwork2012fairness} if the subjects in the protected group and the unprotected group have equal probability of being assigned to the positive predicted class, e.g.~if credit card approval is equally probable for both females and males. 

\begin{figure*}[t]
\centering
\includegraphics[width=0.45\textwidth ]{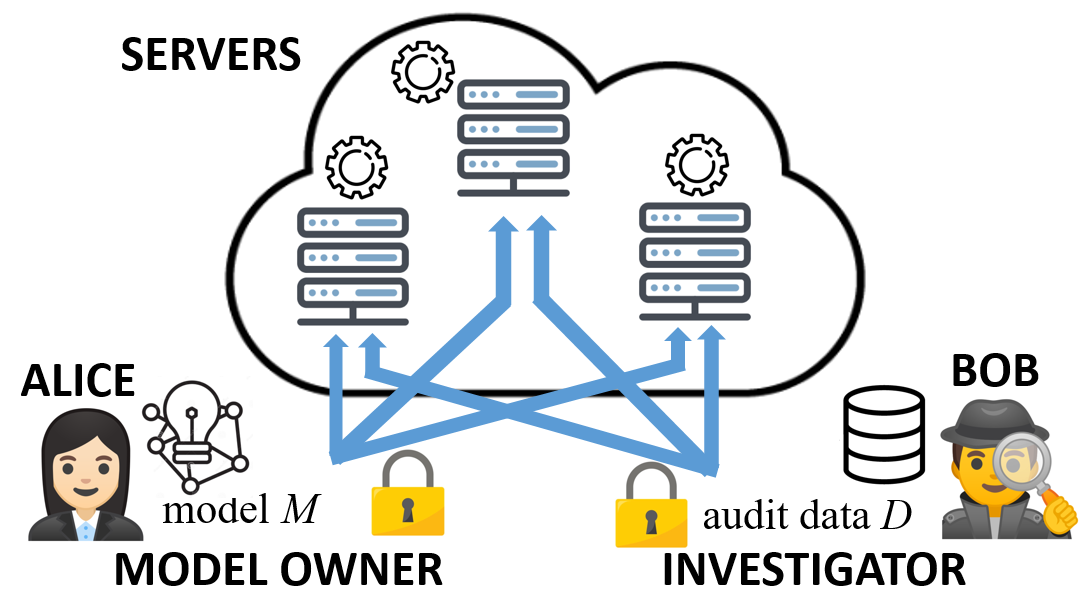}\\
\normalsize{(a) Secure fairness auditing in the 3PC scenario}\\
\, \\
\scriptsize{
\begin{tabular}{|l|c|l|}
\cline{1-1} \cline{3-3}
\texttt{Encrypting model parameters...} 
& \phantom{} & \texttt{Encrypting audit data} \\
\texttt{Connecting to servers...}
& & \texttt{Connecting to servers...}\\
\texttt{Connected to servers...}
& & \texttt{Connected to servers...}\\
\texttt{Secret sharing parameters with servers...}
& & \texttt{Secret sharing audit data with servers...}\\
& & \\
\texttt{Servers running MPC protocol for DP}
& & \texttt{Servers running MPC protocol for DP}\\
\texttt{Servers running MPC protocol for EOP}
& & \texttt{Servers running MPC protocol for EOP}\\
& & \\
\texttt{Finished protocol execution}
& & \texttt{Finished protocol execution}\\
& & \texttt{Aggregating results...}\\
& & \\
& & \texttt{Demographic parity - male: 0.24}\\ 
& & \texttt{Demographic parity - female: 0.19}\\ 
& & \texttt{Equal opportunity - male: 0.399}\\
& & \texttt{Equal opportunity - female: 0.40}\\
\cline{1-1} \cline{3-3}
\end{tabular}
}\\
\, \\
\normalsize{(b) Model owner's terminal during audit} 
\phantom{xxxx}
\normalsize{(c) Investigator's terminal during audit} 
\caption{Alice and Bob send encrypted shares of the model $\mathcal{M}$ (Alice) and the audit data $\mathcal{D}$ (Bob) to 3 servers. The servers subsequently execute MPC protocols to measure demographic parity (DP) and equal opportunity (EOP) in a privacy-preserving manner, i.e.~through computations over the encrypted data.}
\label{fig:screenshot}
\end{figure*}

In real world scenarios, a proprietary model $\mathcal{M}$ held by a company \textit{Alice} may need to be audited by an external investigator \textit{Bob} using sensitive audit data $\mathcal{D}$ (see Figure \ref{fig:screenshot}a).
For example, a bank or hospital, represented by Bob, wants to purchase use of a predictive model and needs to investigate whether the model performs fairly on their data.
%
Alice does not want to disclose her trained model parameters as this could assist rival companies to benefit from her technology. Furthermore, ML models can memorize specific examples from the training data \cite{carlini2019secret},
hence disclosing $\mathcal{M}$, or even giving just black box access through an API interface, can leak very specific information about the training data, which might be sensitive in itself.
Likewise, Bob does not want to disclose the audit data $\mathcal{D}$ to Alice, because it contains sensitive attributes that are on one hand needed for the fairness audit, while on the other hand may be subject of anti-discrimination and data protection law. 

While substantial research progress has been made in the area of fairness in AI and, separately, in the area of privacy-preserving ML, there is a gap in literature that addresses fairness and privacy simultaneously. Existing algorithms for fairness auditing, and their implementations in libraries such as Fairlearn \cite{bird2020fairlearn} and AI Fairness 360 \cite{bellamy2019ai}, operate without concern for privacy, assuming that the entity performing the fairness audit has unrestricted access to the model $\mathcal{M}$ as well as the audit data $\mathcal{D}$.
This assumes that either Alice is willing to disclose $\mathcal{M}$ to Bob, or that Bob is willing to disclose $\mathcal{D}$ to Alice, or that they are both willing to disclose $\mathcal{M}$ and $\mathcal{D}$ to some trusted third party. None of these scenarios is realistic for proprietary models and sensitive audit data.

We present \textsc{PrivFair}, a library for privacy-preserving evaluation of ML models that does not require disclosure of trained model parameters nor of audit data. \textsc{PrivFair} is based on Secure Multiparty Computation (MPC) \cite{damgard}, a cryptographic approach that allows two or more parties to jointly compute a specified output from their private information in a distributed fashion, without actually revealing the private information.

The price paid to preserve privacy during fairness auditing is an increase in runtime, which stems mostly from the computational and communication cost of the MPC protocols used for labeling the audit instances in a privacy-preserving manner.
We argue that this is a reasonable price to pay when working with sensitive data. Indeed, unlike in inference applications -- such as real-time speech recognition or video classification for surveillance -- where fast responsiveness of the system is of utmost importance, in fairness auditing use cases with sensitive data one can typically afford longer runtimes, justifying the use of the most stringent security settings, even if they come at a higher cost.

\section{Preliminaries} 

As is common in MPC, all computations in \textsc{PrivFair} are done over integers in a ring $\mathbb{Z}_q = \{0,1,2,\ldots,q-1\}$. 
The parameter values of Alice's trained model $\mathcal{M}$, and the  feature values of Bob's audit data $\mathcal{D}$, are natively often real numbers represented in a floating point format. 
In this case, as is common in MPC, Alice and Bob convert all 
data in their inputs $\mathcal{M}$ and $\mathcal{D}$ into integers using a bijection between the reals (represented by fixed point notation with $k$ bits in total out of which $a$ bits of decimal precision), and the integers in $\mathbb{Z}_q$, making it simple to go from one representation to the other \cite{drail2019priml}. $q=2^k$ is chosen large enough so that there is no loss of precision that would affect the utility of the ML models.
Next, Alice and Bob encrypt the integers by splitting them into so-called secret shares (see below) and distributing these secret shares across parties (servers) who subsequently perform computations over the shares (see Figure~\ref{fig:screenshot}).
While the original values of the inputs $\mathcal{M}$ and  $\mathcal{D}$ can be trivially revealed by combining all shares, the secret-sharing based MPC schemes ensure that nothing about the inputs is revealed to any subset of the servers that can be corrupted by an adversary. This means, in particular, that none of the servers by themselves learns anything about the actual values of the inputs.

\textsc{PrivFair} incorporates a variety of MPC schemes, designed for different numbers of parties (servers) and offering various levels of security that correspond to different threat models. Regarding threat models, the fairness auditing protocols in \textsc{PrivFair}  provide protection against \textit{passive} as well as  \textit{active} adversaries. While parties corrupted by passive adversaries follow the protocol instructions correctly but try to obtain additional information, parties corrupted by active adversaries can deviate from the protocol instructions. Regarding the number of parties, while fairness auditing is inherently a two-party (2PC) scenario between Alice and Bob, they may not each have a server available on their end and prefer to outsource  computations to the cloud. \textsc{PrivFair}  accommodates two scenarios:

\begin{description}
\item[Scenario 1] \textit{Alice} and \textit{Bob} have the required computational setup to perform MPC computations themselves and proceed by secret-sharing the data with each other only (2PC), 
\item[Scenario 2] \textit{Alice} and \textit{Bob} need to perform computation intensive tasks and outsource the MPC based computations, thus secret-sharing the data across a set of untrusted servers (see Figure \ref{fig:screenshot}a for an illustration of 3PC).
\end{description}

\noindent
The 2PC setting is a \textit{dishonest-majority} setting in which each server can only trust itself. In the 3PC setting, we consider an \textit{honest-majority} setting where at most one of the servers is corrupted by an adversary.

The fairness auditing protocols in \textsc{PrivFair} are sufficiently generic to be used in dishonest-majority as well as honest-majority settings, with passive or active adversaries. This is achieved by changing the underlying MPC scheme to align with the desired security setting. We refer to the caption of Table \ref{tab:mpcresult} for a listing of the various MPC schemes that we used to obtain the results reported in this paper.

In a replicated secret sharing scheme with 3 servers (3PC) and passive adversaries for example \cite{araki2016high}, a value $x$ in $\mathbb{Z}_q$ is secret shared among the servers (parties) $S_1, S_2,$ and $S_3$ by picking uniformly random numbers (shares) $x_1, x_2, x_3 \in \mathbb{Z}_q$ such that 
$x_1 + x_2 +x_3 =  x \mod{q}$,
and distributing $(x_1,x_2)$ to $S_1$, $(x_2,x_3)$ to $S_2$, and $(x_3,x_1)$ to $S_3$. Note that, while the secret-shared information can be trivially revealed by combining all shares, no single server can obtain any information about $x$ given its shares. In the remainder of this paper, we use $[\![x]\!]$ as a shorthand for a secret sharing of $x$, regardless of which secret-sharing scheme is used. In the replicated secret-sharing scheme sketched above, $[\![x]\!] = ((x_1,x_2),(x_2,x_3),(x_3,x_1))$.
%

The MPC schemes used in \textsc{PrivFair}  provide a mechanism for the servers to perform primitive operations over the secret shares, namely addition of a constant, multiplication by a constant, and addition and multiplication of secret-shared values. Building on these cryptographic primitives, MPC protocols for other operations have been developed in the literature. As we explain below, \textsc{PrivFair}'s MPC protocols for fairness auditing rely on established MPC subprotocols for multiplication $\pmul$, 
division $\pdiv$, and equality testing $\peq$ of secret-shared values; see e.g.~\cite{catrina2010secure, dalskov2019secure,dalskov2021fantastic} and references therein for descriptions of these subprotocols.

\section{Secure Fairness Auditing Protocols} 
Below we present MPC protocols for fairness auditing of ML models.
By fairness, we mean algorithmic fairness as commonly understood in the ML literature, namely that a person's experience with an information system should not irrelevantly depend on their personal characteristics such as race, gender, sexual orientation, ethnicity, religion, or age \cite{ekstrand2018privacy}. Fairness has been operationalized through various measures of group fairness, expressing that different groups of users should receive similar statistical treatment. Many different notions of group fairness have been proposed in the literature, some of them even mathematically incompatible \cite{kleinberg2016inherent}; the discussion on the most appropriate metrics is still ongoing and out of scope for this paper.

To the best of our knowledge, the use of MPC for private detection and
mitigation of bias in ML models has so far only been considered by
Kilbertus et al.~\cite{kilbertus80blind}, for one group fairness measure
(statistical parity). Kilbertus et al.~focus on PP training of a fair
logistic regression (LR) model, in an honest-but-curious 2-server set-up
(passive security) in which one of the servers learns the audit data
in-the-clear. The main differences with our project is that we propose
(1) fairness auditing of a variety of ML models (LR, SVM, convolutional
networks (CNNs), decision trees (DTs), and random forests (RFs)), (2)
against a variety of both group and individual fairness measures, (3)
under passive \textit{and} active security threat models, in 2- and
3-server setups, and (4) while fully protecting both the audit data
\textit{and} the model parameters. Our protocols can be applied in tandem
with the technique proposed by Segal et al.~\cite{segal2020fairness} for
certifying that an audited ML model is fair; in contrast with Segal et
al.~we propose MPC-based protocols for auditing the ML models in a PP
way, with audit data that is not disclosed in an unencrypted manner to
the audit servers.

Many well adopted statistical notions of fairness rely on measuring the number of true positives (TP), false positives (FP), true negatives (TN), and false negatives (FN) when applying the ML model under investigation to an audit data set. Privacy-preserving computations of these numbers form the basic building blocks of our proposed protocols. In the description of the protocols below, the audit data $\mathcal{D}$ contains samples of the form $(\vecx,y,a)$ in which $\vecx$ is the instance that needs to be classified (e.g.~an image, or a row of tabular data), $y$ is the ground truth class label, and $a$ indicates whether the instance belongs to an unprivileged demographic subgroup. $a$ is a value of a binary variable $A$ where $A=0$ designates the unprotected group and $A=1$ designates the protected (or sensitive) group. $y$ is a value of a variable $Y$ that represents the actual outcome. Similarly, we use $\hat{Y}$ to denote a variable that represents the predicted outcome according to the model $\mathcal{M}$. For ease of explanation, we first assume that $Y$ and $\hat{Y}$ are binary, i.e.~that we are solving a binary classification task. Next, we explain how the protocols can be used for auditing of multi-class classifiers as well. 


\begin{myprotocol}
   \caption{Protocol $\peqodd$ for computing equalized odds for multi-class classification
   }
   \label{prot:evalfair}
    \textbf{Input:} The parties have a secret sharing of trained model parameters  $[\![\mathcal{M}]\!]$, and a secret sharing of 
    a data set $[\![\mathcal{D}]\!]$ with $N$ instances and secret sharings $[\![Y]\!]$ and $[\![A]\!]$ of the corresponding ground truth labels (drawn from a set $\mathcal{L}$ with $C$ labels) and a binary sensitive attribute. 
    
    \textbf{Output:} A secret sharing of the equalized odds metrics for each class
    
\begin{algorithmic}[1]
   \STATE $[\![Y_\mathsf{pred}]\!]$ $\leftarrow$ $\pinfer([\![\mathcal{M}]\!]$,$[\![\mathcal{D}]\!]$)\\

   \FOR{$c$ = $0$ {\bfseries to} $C-1$} {
   
      \FOR {$i$ = $1$ {\bfseries to} $N$} {
        \STATE  [\![grnd]\!] $\leftarrow$   $\peq([\![Y[i] ]\!]$, $c$)
        
        \STATE  [\![pred]\!] $\leftarrow$   $\peq([\![Y_\mathsf{pred}[i]]\!]$, $c$)
        
        \STATE  [\![a]\!] $\leftarrow$ [\![$A[i]$]\!]

        \STATE  [\![tp]\!] $\leftarrow$ $\pmul$([\![grnd]\!], [\![pred]\!]) 
        \STATE [\![ta]\!] $\leftarrow$ $\pmul$([\![grnd]\!],  [\![a]\!]) 
        \STATE  [\![pa]\!] $\leftarrow$ $\pmul$([\![pred]\!], [\![a]\!])  
        \STATE [\![tpa]\!] $\leftarrow$ $\pmul$([\![tp]\!], [\![a]\!]) 

        \STATE [\![TP$_\mathsf{A=1}[c]$]\!] $\leftarrow$ [\![TP$_\mathsf{A=1}[c]$]\!] +  [\![tpa]\!]
        \STATE [\![FN$_\mathsf{A=1}[c]$]\!] $\leftarrow$ [\![FN$_\mathsf{A=1}[c]$]\!] + ([\![ta]\!] $-$ [\![tpa]\!])
        \STATE [\![FP$_\mathsf{A=1}[c]$]\!] $\leftarrow$ [\![FP$_\mathsf{A=1}[c]$]\!] +  ([\![pa]\!] $-$ [\![tpa]\!])
        \STATE [\![TN$_\mathsf{A=1}[c]$]\!] $\leftarrow$ [\![TN$_\mathsf{A=1}[c]$]\!] \\
        \phantom{[\![TN$_\mathsf{A=1}[c]$]\!] $\leftarrow$} +  ([\![a]\!] $-$ [\![ta]\!] $-$ [\![pa]\!] + [\![tpa]\!])
        
        \STATE  [\![TP$_\mathsf{A=0}[c]$]\!] $\leftarrow$ [\![TP$_\mathsf{A=0}[c]$]\!] + ([\![tp]\!] $-$ [\![tpa]\!])
         
        \STATE  [\![FN$_\mathsf{A=0}[c]$]\!] $\leftarrow$ [\![FN$_\mathsf{A=0}[c]$]\!]\\
        \phantom{[\![FN$_\mathsf{A=0}[c]$]\!] $\leftarrow$}
        + ([\![grnd]\!] $-$ [\![ta]\!] $-$ [\![tp]\!] + [\![tpa]\!])
        \STATE [\![FP$_\mathsf{A=0}[c]$]\!] $\leftarrow$ [\![FP$_\mathsf{A=1}[c]$]\!] \\
        \phantom{[\![FP$_\mathsf{A=0}[c]$]\!] $\leftarrow$}+  ([\![pred]\!] $-$ [\![pa]\!] $-$ [\![tp]\!] + [\![tpa]\!])
        \STATE [\![TN$_\mathsf{A=0}[c]$]\!] $\leftarrow$ [\![TN$_\mathsf{A=1}[c]$]\!]\\ 
        \phantom{[\![TN$_\mathsf{A=0}[c]$]\!] $\leftarrow$} +  (1 $-$ [\![grnd]\!] $-$ [\![pred]\!] $-$ [\![a]\!] \\
        \phantom{[\![TN$_\mathsf{A=0}[c]$]\!] $\leftarrow$} + [\![tp]\!] + [\![ta]\!]+ [\![pa]\!] $-$ [\![tpa]\!])

        }
        \ENDFOR
        }
        \STATE [\![TPR$_\mathsf{A=1}[c]]\!]$ $\leftarrow$ $\pdiv($[\![TP$_\mathsf{A=1}[c]$]\!] ,\\ \phantom{[\![TPR$_\mathsf{A=1}[c]]\!]$ $\leftarrow$ $\pdiv($}[\![TP$_\mathsf{A=1}[c]$]\!] + [\![FN$_\mathsf{A=1}[c]$]\!])
        \STATE [\![TPR$_\mathsf{A=0}[c]]\!]$ $\leftarrow$ $\pdiv($[\![TP$_\mathsf{A=0}[c]$]\!] ,\\ \phantom{[\![TPR$_\mathsf{A=0}[c]]\!]$ $\leftarrow$ $\pdiv($}[\![TP$_\mathsf{A=0}[c]$]\!] + [\![FN$_\mathsf{A=0}[c]$]\!])
        
       \STATE [\![FPR$_\mathsf{A=1}[c]]\!]$ $\leftarrow$ $\pdiv($[\![FP$_\mathsf{A=1}[c]$]\!] ,\\ \phantom{[\![FPR$_\mathsf{A=1}[c]]\!]$ $\leftarrow$ $\pdiv($}[\![FP$_\mathsf{A=1}[c]$]\!] + [\![TN$_\mathsf{A=1}[c]$]\!])
        \STATE [\![FPR$_\mathsf{A=0}[c]$]\!] $\leftarrow$ $\pdiv($[\![FP$_\mathsf{A=0}[c]$]\!] ,\\ \phantom{[\![FPR$_\mathsf{A=0}[c]$]\!] $\leftarrow$ $\pdiv($}[\![FP$_\mathsf{A=0}[c]$]\!] + [\![TN$_\mathsf{A=0}[c]$]\!])

        \ENDFOR
   \STATE \textbf{return} [\![TPR$_\mathsf{A|0,1}]\!]$ , [\![FPR$_\mathsf{A|0,1}]\!]$
\end{algorithmic}
\end{myprotocol}

\paragraph{Protocol $\boldpeqodd$ for Equalized Odds} 

By definition, equalized odds is satisfied if $\hat{Y}$ and $A$ are independent conditional on $Y$ (Eq.~(\ref{EOdd})) \cite{hardt2016equality}, i.e.
\begin{equation}\label{EOdd}
\hat{Y} \indep A \ | \ Y
\end{equation}

For a binary classifier, this is equivalent to Eq.~(\ref{EOdd_binary}), implying that the true positive rate TPR = TP/(TP+FN) and the false positive rate FPR = FP/(FP+TN) are the same for both the unprotected and the protected groups.
An exact equality as in Eq.~(\ref{EOdd_binary}) may be hard to achieve, so it is common to compute the left and the right hand sides of Eq.~(\ref{EOdd_binary}) separately and inspect their differences, sometimes reported through metrics such as the equalized odds difference. 
%
\begin{equation}\label{EOdd_binary}
\begin{array}{l}
\mbox{P}(\hat{Y} = 1 \,|\, Y=y,A=0)=\mbox{P}(\hat{Y} = 1 \,|\, Y=y,A=1), \\ 
\forall y \in \{0,1\}
\end{array}
\end{equation}

We extend Eq.~(\ref{EOdd_binary}) to multi-class classification using the \textit{one-vs-rest} approach. For a classification task with a label set $\mathcal{L} = \{0,1,...,C-1\}$, for each class label $c \in \mathcal{L}$, we define the equalized odds for class label $c$ as per Eq.~(\ref{EOdd_multi}). In this case, for each class label $c$, we consider two cases with $\hat{Y} = c$ and $\hat{Y} = \lnot c$ (any other class). 
Adapting Eq.~(\ref{EOdd_binary}) to this gives us 
Eq.~(\ref{EOdd_multi}) which implies that the TPR and FPR for each class should be same for both the protected and unprotected groups.
\begin{equation}\label{EOdd_multi}
\hskip -0.5pt
\begin{array}{l}
\mbox{P}(\hat{Y} = c \,|\, Y=y,A=0)=\mbox{P}(\hat{Y} = c \,|\, Y=y,A=1), \\ 
\forall y \in \{c,\lnot c\}, \forall c \in \mathcal{L}
\end{array}
\end{equation}




In \textsc{PrivFair}, the left and right hand sides of Eq.~(\ref{EOdd_multi}) are computed in a privacy-preserving manner with protocol $\peqodd$, presented in pseudocode here as Protocol \ref{prot:evalfair}. At the beginning of this protocol, the servers (parties) have secret shares of the parameters of the model $\mathcal{M}$, as received from the model owner Alice (see Figure \ref{fig:screenshot}). Similarly, from the investigator Bob, the parties have received secret shares of an audit data set $\mathcal{D}$ with $N$ instances, including a secret-shared vector $Y$ of length $N$ with the ground truth labels and a secret-shared vector $A$ of length $N$ with the sensitive attribute. Note that we abuse the notations $Y$ and $A$ here to denote random variables as in Eq.~(\ref{EOdd_multi}) as well as vectors with values for those random variables as in Protocol \ref{prot:evalfair}.

On Line 1, the servers perform privacy-preserving labeling of the audit data instances in $\mathcal{D}$ with the model $\mathcal{M}$ using an MPC protocol $\pinfer$ for secure inference. Such protocols have been developed by us and others for logistic regression, neural networks, decision tree ensembles etc.~\cite{IEEETDSC:CDHK+17,fritchman2018,agrawal2019quotient,de2019privacy,dalskov2019secure,icmlsikha} and can be used in combination with the protocols in \textsc{PrivFair}. After Line 1, the servers have a secret-shared vector $[\![Y_\mathsf{pred}]\!]$ with predicted class labels for each of the samples in the audit data.

Next, on Line 2--19, for each of the classes, the servers compute the number of true positives (TP),  number of false positives (FP), number of true negatives (TN) and false negatives (FN) for the (un)protected group.
To this end, on Line 4, the servers compute a secret-shared binary variable $[\![$grnd$]\!]$ representing if instance $i$ belongs to the class $c$ that is being inspected. To obtain this, the servers run an MPC protocol $\peq$ for equality testing that takes as input the secret-shared ground truth class label $[\![Y[i]]\!]$ and the class $c$ itself, and returns a secret-sharing of $1$ if the equality test was positive, and a secret-sharing of $0$ otherwise.
Next, on Line 5, the servers compute in a similar way a secret-shared binary variable $[\![$pred$]\!]$ representing if the predicted outcome for instance $i$ is the class $c$ that is being inspected. On Line 6, the servers compute a secret-shared binary value $[\![$a$]\!]$ representing if the current sample in the audit data belongs to the protected or unprotected group.  

\begin{table}
\centering
\begin{tabular}{|l|l|l|l|l|l|}
\hline
grnd & pred  & Metric  & ~~~~~~Logic \\ 
\hline
~~1 & ~~1 & ~~TP  & ~~ grnd $*$ pred \\ 
~~1 & ~~0 & ~~FN  & ~~ grnd $*$ ($1-$pred) \\ 
~~0 & ~~1 & ~~FP & ~~ (1-grnd) $*$ pred \\ 
~~0 & ~~0 & ~~TN  & ~~ (1-grnd) $*$ ($1-$pred) \\
\hline
\end{tabular}
\caption{Logic for evaluating TP, FN, FP and TN} 
\label{tab:truthtable}
\end{table}

The underlying logic for efficiently computing TP, FN, FP, and TN, is derived from the truth table shown in Table \ref{tab:truthtable}. 
Multiplying column 4 in Table \ref{tab:truthtable} with a and with $1-$a respectively gives us the contribution of instance $i$ to the TP, FN, FP, and TN metrics for the protected and unprotected group respectively. For example, the total number of instances of the protected group that get incorrectly classified as belonging to class $c$ is
\begin{equation}\label{sumFP}
\mbox{FP}_{\mbox{\small A}=1}[c] = \sum\limits_{i=1}^n (1-\mbox{grnd}^{(i)}) * \mbox{pred}^{(i)} * \mbox{a}^{(i)}   
\end{equation}
in which $\mbox{grnd}^{(i)}$, $\mbox{pred}^{(i)}$, and $\mbox{a}^{(i)}$ are the secret-shared binary variables computed for instance $i$ on Line 4--5.

This logic is modified to suit the MPC computations on Line 7--18. To this end, we rewrote the expressions in the summations as in the right hand side of Eq.~(\ref{sumFP}) to reduce the overall number of multiplications as much as possible. As a result, $\peqodd$ requires only 4 multiplications per instance and per class (see Lines 7--10). 
On Lines 11--14 and Lines 15--18, the servers use these precomputed secret-shared products to obtain secret shares of TP, FN, FP, and TN respectively for the protected group and for the unprotected group. 

Finally, on Line 20--23 the servers execute a protocol $\pdiv$ for division with secret-shared values, to obtain secret shares of the TPR and FPR for each class for the protected and unprotected groups. 
At the end of the protocol, each server sends its secret shares to the investigator Bob who can then combine the shares to learn the TPR and FPR values. The servers themselves do not learn anything about the real values of the TPR and the FPR, nor of the TP, FN, FP, and TN metrics computed along the way, and not even the number of instances of each class in the audit data. 

While Protocol \ref{prot:evalfair} is presented for multi-class classification, it can be trivially used for binary classification as well by removing Line 2 and 19, and running the inner for-loop only once with a fixed value of $c=1$.


\paragraph{Protocol $\boldpeqopp$ for Equal Opportunity} 
By definition, equal opportunity for a binary predictor is satisfied if
\cite{hardt2016equality} 
\begin{equation}\label{EOpp}
\hskip -0.5pt
\mbox{P}(\hat{Y} = 1 \,|\, Y=1,A=0) = \mbox{P}(\hat{Y} = 1 \,|\, Y=1,A=1)
\end{equation}
implying that the TPR is the same for both the unprotected and the protected group. This fairness notion is a relaxation of equalized odds in a sense that it focuses only the positive or advantaged outcome.

Protocol \ref{prot:evalfair} can be modified directly to compute Eq.~(\ref{EOpp}) resulting in a new protocol $\peqopp$. This can be done by running the inner for-loop only once, with a fixed value of $c=1$ (binary classification) and by removing several lines, i.e.~Line 9; the lines that compute FP and TN (Lines 13--14,17--18); and the lines that compute FPR (Lines 22--23). Through executing the resulting protocol $\peqopp$, the servers will only compute secret shares of TPR (Line 25) for $c=1$.

\begin{myprotocol}
   \caption{Protocol $\pacc$ for computing sub-group accuracy for multi-class classification
   }
   \label{prot:acc}
    \textbf{Input:} The parties have a secret sharing of trained model parameters  $[\![\mathcal{M}]\!]$, and a secret sharing of 
    a data set $[\![\mathcal{D}]\!]$ with $N$ instances and secret sharings $[\![Y]\!]$ and $[\![A]\!]$ 
    of the corresponding ground truth labels (drawn from a set $\mathcal{L}$ with $C$ labels) and a binary sensitive attribute. 
    
    \textbf{Output:} A secret sharing of the sub-group accuracy metrics
    
\begin{algorithmic}[1]

    \STATE $[\![Y_\mathsf{pred}]\!]$ $\leftarrow$ $\pinfer([\![\mathcal{M}]\!]$,$[\![\mathcal{D}]\!]$)\\
   
    \FOR {$i$ = $1$ {\bfseries to} $N$} {
    \STATE [\![count$_\mathsf{A=1}$]\!] $\leftarrow$ [\![count$_\mathsf{A=1}$]\!] + [\![$A[i]$]\!]
    }\ENDFOR
    
    \STATE [\![count$_\mathsf{A=0}$]\!] $\leftarrow$ $N$ $-$ [\![count$_\mathsf{A=1}$]\!]
    
    \STATE [\![correct$_\mathsf{A=1}$]\!] $\leftarrow$  $0$
    \STATE [\![correct$_\mathsf{A=0}$]\!] $\leftarrow$  $0$
    
    

    \FOR {$i$ = $1$ {\bfseries to} $N$} {
    \STATE [\![corr]\!] $\leftarrow$ [\![$Y[i]$]\!] $-$ [\![$Y_\mathsf{pred}[i]$]\!]
    
    \STATE [\![iscorr]\!] $\leftarrow$ $\peq$([\![corr]\!], $0$)
    \STATE [\![iscorr$_\mathsf{A=1}$]\!] $\leftarrow$ $\pmul$( [\![iscorr]\!],  [\![$A[i]$]\!])
    
   \STATE [\![correct$_\mathsf{A=1}$]\!] $\leftarrow$ [\![correct$_\mathsf{A=1}$]\!] + [\![iscorr$_\mathsf{A=1}$]\!]
    
   \STATE [\![correct$_\mathsf{A=0}$]\!] $\leftarrow$ [\![correct$_\mathsf{A=0}$]\!] \\
   \phantom{[\![correct$_\mathsf{A=0}$]\!] $\leftarrow$}+ ([\![iscorr]\!] $-$ [\![iscorr$_\mathsf{A=1}$]\!])
    
    }\ENDFOR

    \STATE [\![ACC$_\mathsf{A=1}$]\!] $\leftarrow$ $\pdiv$([\![correct$_\mathsf{A=1}$]\!] , [\![count$_\mathsf{A=1}$]\!])
    

    \STATE [\![ACC$_\mathsf{A=0}$]\!] $\leftarrow$ $\pdiv$([\![correct$_\mathsf{A=0}$]\!] , [\![count$_\mathsf{A=0}$]\!]) 
    

    \STATE [\![ACC]\!] $\leftarrow$ $\pmul$([\![correct$_\mathsf{A=1}$]\!] + [\![correct$_\mathsf{A=0}$]\!], 1/$N$)

   \STATE \textbf{return} [\![ACC$_\mathsf{A|0,1}]\!]$, [\![ACC]\!]
\end{algorithmic}
\end{myprotocol}

\begin{table*}
\centering

\resizebox{\linewidth}{!}{

\begin{tabular}{l|c|c|lll|ll}
    ~ ~ ~ \textbf{Task}   & 
    \begin{tabular}[c]{@{}l@{}}\textbf{\#Samples}\\(owned \\by Bob)\end{tabular} & 
    \begin{tabular}[c]{@{}l@{}}\textbf{\#Model} \\\textbf{Parameters}\\(owned \\by Alice)\end{tabular} &
        & 
    \multicolumn{2}{l|}{~ ~ ~ ~ ~ ~ ~ ~\textbf{Passive}}  & 
    \multicolumn{2}{l}{~ ~ ~ ~ ~ ~ ~ ~ ~ ~ ~ ~ ~\textbf{Active}}                                                     \\ 
\hline
    &    
    &
    &     
    & 
    \begin{tabular}[c]{@{}l@{}}DP (\pdp)\end{tabular} &
    \begin{tabular}[c]{@{}l@{}}EOP (\peqopp)\end{tabular} & 
    \begin{tabular}[c]{@{}l@{}}DP (\pdp)\end{tabular} & 
    \begin{tabular}[c]{@{}l@{}}EOP (\peqopp)\end{tabular} 
    \\ 
\cline{5-8}

\begin{tabular}[c]{@{}l@{}}Binary \\classification\end{tabular} & 
~ ~200  & 
\begin{tabular}[c]{@{}l@{}}~ 47 \\~(LR)\end{tabular} & 
2PC &  
  \phantom{x,xx}3.34 sec &                                                                   
  \phantom{x,xxx}3.36 sec &                                                                    
  ~ ~  239.24 sec &  
  \phantom{xx,}238.69 sec 
\\
  &  
  & 
  & 
  3PC & 
  \phantom{x,xx}1.37 sec &                                                                   
  \phantom{x,xxx}1.67 sec &                                                                    
  ~ ~  \phantom{xx}6.41 sec &  
   \phantom{xx,xx}6.43 sec  
  \\ 
\hline
   & 
   & 
   &     
   & 
   \begin{tabular}[c]{@{}l@{}}EOD (\peqodd)\end{tabular}   & 
   \begin{tabular}[c]{@{}l@{}}Grp. Acc (\pacc)\end{tabular}  & 
   \begin{tabular}[c]{@{}l@{}}EOD (\peqodd)\end{tabular}  & 
   \begin{tabular}[c]{@{}l@{}}Grp. Acc (\pacc)\end{tabular}
   \\ 
\cline{5-8}
\begin{tabular}[c]{@{}l@{}}Multi-class \\classification\end{tabular} & 
~ ~ ~56  & 
\begin{tabular}[c]{@{}l@{}}~ ~ $1.48$M \\~(ConvNet)\end{tabular} & 
2PC &  
   5,866.47 sec &                                                                   
~   5,866.46 sec &                                                                    
  42,745.09 sec  &  
  42,742.00 sec 
\\
                                                                                                  
 &                                                                        
 &                                                                                 
 & 
  3PC & 
 \phantom{x,x}29.92 sec & 
 ~ \phantom{x,x}29.84 sec & 
   \phantom{xx,}199.02 sec & 
   \phantom{xx,}198.99 sec 
   
\end{tabular}

}
\caption{Time taken to execute individual MPC protocols in \textsc{PrivFair} -- includes time for making inferences and fairness evaluation. Binary classification corresponds to credit score classification on the German credit score data set. Multi-class classification corresponds to detecting one of 7 emotions in an image using the RAVDESS data set. The times do not include compile time and are an average over 5 runs. DP: Demographic Parity, EOP: Equal Oppurtunity, EOD: Equalized Odds, Grp. Acc: Sub-group accuracy. 2PC-Passive: \cite{cryptoeprint:2018:482}, 2PC-Active: \cite{damgaard2019new} , 3PC-Passive: \cite{araki2016high} , 3PC-Active: \cite{dalskov2021fantastic}.}
\label{tab:mpcresult}

\end{table*}

\paragraph{Protocol $\boldpacc$ for Sub-Group Accuracy} 
This notion of fairness, similar to overall accuracy equality \cite{berk2021fairness}, is satisfied if the classifier is equally accurate for both the protected and unprotected group.

To report sub-group accuracy, \textsc{PrivFair}'s protocol $\pacc$ (see Protocol \ref{prot:acc}) computes the accuracy for the protected group, the accuracy for the unprotected group, and the overall accuracy. These notions are defined as follows, for a classifier $\mathcal{M}$ and an audit data set $\mathcal{D}$ with instances of the form $(\vecx,y,a)$:
\begin{equation}\label{ACC1}
    \mbox{ACC}_\mathsf{A=1} = \frac{|\{(\vecx,y,1) \in \mathcal{D} \,|\, y = \mathcal{M}(\vecx)\}|}{|\{(\vecx,y,1) \in \mathcal{D}\}|}
\end{equation}
\begin{equation}\label{ACC0}
    \mbox{ACC}_\mathsf{A=0} = \frac{|\{(\vecx,y,0) \in \mathcal{D} \,|\, y = \mathcal{M}(\vecx)\}|}{|\{(\vecx,y,0) \in \mathcal{D}\}|}
\end{equation}
\begin{equation}\label{ACCAll}
    \mbox{ACC} = \frac{|\{(\vecx,y,a) \in \mathcal{D} \,|\, y = \mathcal{M}(\vecx)\}|}{|\{(\vecx,y,a) \in \mathcal{D}\}|}
\end{equation}
Eq.~(\ref{ACC1})--(\ref{ACCAll}) cover both binary and multi-class classifiers.

On Line 1 in Protocol \ref{prot:acc}, the servers perform privacy-preserving labeling of the audit data $\mathcal{D}$ with the model $\mathcal{M}$. On Line 2-4, the servers ``count'' how many instances in the audit data belong to the protected group, i.e.~the denominator of Eq.~(\ref{ACC1}). Note that each of the servers obtains only a secret share of the resulting
$\mbox{count}_\mathsf{A=1}$, i.e.~no server finds out how many instances in the audit data belong to the protected class. Since the total number of instances $N$ is public information, the servers can straightforwardly retrieve a secret-sharing of the number of instances belonging to the unprotected group by computing $N - [\![\mbox{count}_\mathsf{A=1}]\!]$. This is leveraged in Line 5.  


Through Lines 8--14, the servers count how many predictions are correct. To this end, for each instance, on Line 9 the servers compute the difference between the actual outcome and predicted outcome. If this difference is equal to zero as per Line 10, it means it was a correct prediction. Lines 11--13 decide whether this correct prediction is to be accounted for in the total of the protected or the unprotected group. The right hand side of Line 13 for the unprotected group is written to take advantage of the multiplication that was already done for the protected group on Line 11.

 
On Line 15--16 the servers compute accuracy for the protected and unprotected group respectively. Line 17 computes overall accuracy by performing secure addition of number of correct instances for each subgroup followed by secure multiplication with a constant $1/N$.

\paragraph{Protocol $\boldpdp$ for Demographic Parity} 
Demographic parity, also known as statistical parity, is one of the best known and most widely accepted notions of group fairness. A classifier satisfies demographic parity if \cite{dwork2012fairness}  
\begin{equation} \label{dp}
\hskip -0.5pt
\mbox{P}(\hat{Y} = 1 \,|\, A=0)  =   \mbox{P}(\hat{Y} = 1 \,|\, A=1)  
\end{equation}
implying that both protected and unprotected groups have equal probability of receiving positive outcomes \cite{verma2018fairness}. This is equivalent to the ratio of the number of positive outcomes for the group to the total number of instances belonging to the group. Note that the number of positive outcomes for a group in the audit data is the sum of the number of true positives and the number of false positives for that group, an idea that is leveraged in \textsc{PrivFair}'s MPC protocol $\pdp$ for demographic parity.



Protocol $\pdp$ can be derived by replacing Line 2 in Protocol \ref{prot:evalfair} with $c=1$ (for binary classification) and retaining only the lines required for computing TP and FP for each subgroup (Lines 3--10,11,13,15,17). Furthermore, Lines 20-23 are replaced by lines that compute the sum of TP and FP:

\begin{center}
\begin{tabular}{lll}
$[\![$POS$_{\mathsf{A=1}}]\!]$ & $\leftarrow$ & $[\![$TP$_{\mathsf{A=1}}]\!] +   [\![$FP$_{\mathsf{A=1}}]\!]$\\
$[\![$POS$_{\mathsf{A=0}}]\!]$ & $\leftarrow$ & $[\![$TP$_{\mathsf{A=0}}]\!] +   [\![$FP$_{\mathsf{A=0}}]\!]$ \end{tabular}
\end{center}
Additionally to compute the ratio, we adopt computations similar to Protocol \ref{prot:acc}, by including lines for computing counts of the instances belonging to each group (Lines 2--5), followed by the lines to compute the ratio


\begin{center}
\begin{tabular}{lll}

$[\![$DP$_{\mathsf{A=1}}]\!]$ & 
$\leftarrow$ &

$\pdiv$([\![POS$_\mathsf{A=1}$]\!] , [\![count$_\mathsf{A=1}$]\!]) 

 \\

$[\![$DP$_{\mathsf{A=0}}]\!]$ & 
$\leftarrow$ & 
$\pdiv$([\![POS$_\mathsf{A=0}$]\!] , [\![count$_\mathsf{A=0}$]\!]) 
\end{tabular}
\end{center}

At the end, Line 25 in Protocol \ref{prot:evalfair} is replaced with
\begin{center}
\textbf{return} [\![DP$_\mathsf{A=1}$]\!],  [\![DP$_\mathsf{A=0}$]\!]
\end{center}

\section{Results}


\textsc{PrivFair} is implemented\footnote{\url{https://bitbucket.org/uwtppml/privfair}} on top of MP-SPDZ \cite{cryptoeprint:2020:521} and is sufficiently generic to take advantage of the variety of underlying MPC schemes in MP-SPDZ. 

We perform all our experiments on virtual machines (host) on Google Cloud Platform (GCP) with 8 vCPUs, 32 GB RAM and egress bandwidth limited to 16 Gbps. 
We use mixed computations \cite{cryptoeprint:2018:403,cryptoeprint:2018:762,demmler2015aby, cryptoeprint:2020:338} that switch between arithmetic ($\mathbb{Z}_{2^k}$ with $k=64$) and binary ($\mathbb{Z}_{2}$) computations for efficiency. 
All integer additions and multiplications are performed over the arithmetic domain and any non-linear functions such as comparisons are computed over the binary domain.

We report the time taken to execute the MPC protocols in \textsc{PrivFair} under different security settings in Table \ref{tab:mpcresult}, namely for 2PC and 3PC with passive or with active adversaries. To perform experiments for \textit{binary classification}, we audit a logistic regression (LR) model for credit score classification on the benchmark German credit score data set \cite{Dua:2019}. The servers run MPC protocols $\pdp$ and $\peqopp$ from \textsc{PrivFair} to compute demographic parity and equality of opportunity, using gender as the sensitive attribute. For \textit{multi-class classification}, we demonstrate \textsc{PrivFair} to audit a ConvNet ($\sim1.48$M parameters) model for emotion recognition from images \cite{icmlsikha}.
As audit data, we use 56 images from the RAVDESS data set \cite{livingstone2018ryerson} corresponding to different emotions, namely \textit{neutral}, \textit{happy}, \textit{sad},  \textit{angry}, \textit{fearful}, \textit{disgust}, and \textit{surprised}. Using gender as the sensitive attribute, the servers run MPC protocols $\peqodd$ and $\pacc$ from \textsc{PrivFair} to compute equalized odds  and subgroup accuracy.

Regarding utility we note that the LR and ConvNet models were trained in the clear, and that the $\pinfer$ protocol for secure inference with these models infer the same labels as one would obtain without encryption, i.e.~there is no loss of utility (accuracy). The results of the secure fairness auditing protocols are therefore also the same as one would obtain without encryption. In the remainder of this section, we therefore focus on the runtimes.

The times reported in Table \ref{tab:mpcresult} correspond to the entire runtime of the MPC protocols, i.e.~both the so-called offline and online phases. The offline phase includes any preprocessing required to begin executing the MPC protocol (such as the generation of the correlated randomness that is needed for the secure multiplication $\pmul$) and is independent of the specific input values; the online phase is where the MPC protocol executes on the specific inputs. 

Furthermore, the times reported in Table \ref{tab:mpcresult} encompass the time needed to infer class labels for all the instances in the audit data (such as on Line 1 in Protocol \ref{prot:evalfair} and Protocol \ref{prot:acc}) and the time needed to evaluate the individual applicable fairness notions. The substantial differences in runtime for auditing the logistic regression model (binary classification) vs.~the ConvNet model (multi-class classification) stem from large differences in runtime for the inference step. 
For example, auditing 56 images belonging to 7 classes with a ConvNet of 1.48M parameters to report EOD takes 42.745.09 sec in the active 2PC setting. Out of this, 42515.82 sec were taken to classify the 56 images in the audit data in a privacy-preserving manner (Line 1 of Protocol \ref{prot:evalfair}), while execution of the rest of 
Protocol \ref{prot:evalfair} to evaluate the fairness metric itself took 229.27 sec. Nearly all the time taken by 
\textsc{PrivFair}'s protocols for auditing the image classifier are spent on inference, while the time taken to evaluate only the fairness metrics given the labels is near real-time and practical. 

Our runtime results for the time-consuming inference (the first step in the fairness auditing protocols) align with the observations in the MPC literature \cite{dalskov2019secure,dalskov2021fantastic,icmlsikha} where there is also a substantial difference between protocols for passive and active adversarial settings. 
Providing security in the presence of active or \textit{``malicious''} adversaries, i.e.~ensuring that no such adversarial attack can succeed, comes at a much higher computational cost than in the passive case. The same holds for a dishonest-majority 2PC setting where each party (server) only trusts itself, versus a much faster 3PC honest-majority setting. We note that unlike in privacy-preserving inference applications, where fast responsiveness of the system can be of utmost importance for practical applications \cite{dalskov2019secure,icmlsikha}, in fairness auditing use cases such as the ones we consider in this paper, one can typically afford longer runtimes, justifying the use of the most stringent security settings.

All the above results carry over to any similar, same sized audit data sets and models. With increase in the size of the data set or the number of parameters of the model, the times for execution of the MPC protocols will increase linearly \cite{icmlsikha}.

\section{Conclusion \& Future Directions}
In this paper, we presented \textsc{PrivFair}, a first-of-its-kind library for privacy-preserving fairness audits of ML models. The protocols in \textsc{PrivFair} allow an investigator Bob with audit data  to evaluate the fairness of a model owner Alice's ML model without requiring Bob or Alice to disclose their inputs to anyone in an unencrypted manner. The computations required for the fairness audits are performed by servers (parties) in a privacy-preserving manner. To protect both the trained model parameters and the audit data, we use Secure Multi-Party Computation (MPC) which allows for parties to jointly compute functions over encrypted shares of data. \textsc{PrivFair} enables 2PC scenarios in which Alice and Bob act as the 2 parties themselves, as well as 3PC scenarios in which Alice and Bob outsource the fairness audit to untrusted servers in the cloud. The latter service can be adapted to be used not only for synchronous but also asynchronous MPC computations, where servers each store the corresponding  secret shares of the models that are to be investigated and an external investigator can choose the desired model to investigate.

The current open-source implementation of \textsc{PrivFair} includes MPC protocols for the well established fairness metrics of equalized odds, equal opportunity, subgroup accuracy, and demographic parity. 
The fairness auditing protocols can be selected based on their suitability to the application where fairness is to be guaranteed. \textsc{PrivFair} can be easily used and extended to report other statistical notions of fairness \cite{verma2018fairness} such as treatment equality \cite{berk2021fairness} 
and predictive equality \cite{corbett2017algorithmic}. 
Furthermore, while we have demonstrated the use of \textsc{PrivFair} for auditing of logistic regression and ConvNet models, the fairness auditing protocols can be used for any kind of model architecture for which an MPC protocol for secure inference is available, including decision tree models, random forests, support vector machines, and other kinds of neural networks~\cite{IEEETDSC:CDHK+17,fritchman2018,agrawal2019quotient,de2019privacy,dalskov2019secure}.

The applicability and usefulness of \textsc{PrivFair} stretches well beyond the applications considered in our experiments. AI services that deal with sensitive data such as in healthcare, banking, predictive policing etc.~not only need to protect data but also ensure that unbiased AI services are available to all. \textsc{PrivFair} is a first step where AI developers and AI end users can collaborate to aim for unbiased AI services, while protecting their data. As such, the impact of using \textsc{PrivFair} can be significant in all fields where ML models appear for automated decision making, e.g., education, housing, law-enforcement, healthcare, and banking, as well as new application domains yet to be discovered.

\section{Acknowledgements} 
Funding support for project activities has been partially provided by The University of Washington Tacoma’s Founders Endowment fund, Canada CIFAR AI Chair, Facebook Research Award for Privacy Enhancing Technologies, and the Google Cloud Research Credits Program.

\bibliography{references2}

\end{document}